# Multisensor Images Fusion Based on Feature-Level

Firouz Abdullah Al-Wassai[1]
Research Student, Computer Science Dept.
(SRTMU), Nanded, India
fairozwaseai@yahoo.com

N.V. Kalyankar[2]
Principal, Yeshwant Mahavidyala College
Nanded, India
drkalyankarnv@yahoo.com

Ali A. Al-Zaky[3]
Assistant Professor, Dept. of Physics, College of Science, Mustansiriyah Un.
Baghdad – Iraq.
dr.alialzuky@yahoo.com

*Abstract :* Until now, of highest relevance for remote sensing data processing and analysis have been techniques for pixel level image fusion. So, This paper attempts to undertake the study of Feature-Level based image fusion. For this purpose, feature based fusion techniques, which are usually based on empirical or heuristic rules, are employed. Hence, in this paper we consider feature extraction (FE) for fusion. It aims at finding a transformation of the original space that would produce such new features, which preserve or improve as much as possible. This study introduces three different types of Image fusion techniques including Principal Component Analysis based Feature Fusion (PCA), Segment Fusion (SF) and Edge fusion (EF). This paper also devotes to concentrate on the analytical techniques for evaluating the quality of image fusion (F) by using various methods including (SD), (En), (CC), (SNR), (NRMSE) and (DI) to estimate the quality and degree of information improvement of a fused image quantitatively.

## I. INTRODUCTION

Over the last years, image fusion techniques have interest within the remote sensing community. The reason of this is that in most cases the new generation of remote sensors with very high spatial resolution acquires image datasets in two separate modes: the highest spatial resolution is obtained for panchromatic images (PAN) whereas multispectral information (MS) is associated with lower spatial resolution [1].

Usually, the term 'fusion' gets several words to appear, such as merging, combination, synergy, integration … and several others that express more or less the same meaning the concept have since it appeared in literature [Wald L., 1999a]. Different definitions of data fusion can be found in literature, each author interprets this term differently depending on his research interests, such as [2, 3] . A general definition of data fusion can be adopted as fallowing: "Data fusion is a formal framework which expresses means and tools for the alliance of data originating from different sources. It aims at obtaining information of greater quality; the exact definition of 'greater quality' will depend upon the application" [4-6].

Image fusion techniques can be classified into three categories depending on the stage at which fusion takes place; it is often divided into three levels, namely: pixel level, feature level and decision level of representation [7,8]. Until now, of highest relevance for remote sensing data processing and analysis have been techniques for pixel level image fusion for which many different methods have been developed and a rich theory exists [1]. Researchers have shown that fusion techniques that operate on such features in the transform domain yield subjectively better fused images than pixel based techniques [9].

For this purpose, feature based fusion techniques that is usually based on empirical or heuristic rules is employed. Because a general theory is lacking fusion, algorithms are usually developed for certain applications and datasets. [10]. In this paper we consider feature extraction (FE) for fusion.It is aimed at finding a transformation of the original space that would produce such new features, which are preserved or improved as much as possible. This study introduces three different types of Image fusion techniques including Principal Component Analysis based Feature Fusion (PCA), Segment Fusion (SF) and Edge fusion (EF). It will examine and estimate the quality and degree of information improvement of a fused image quantitatively and the ability of this fused image to preserve the spectral integrity of the original image by fusing different sensor with different characteristics of temporal, spatial, radiometric and Spectral resolutions of TM & IRS-1C PAN images. The subsequent sections of this paper are organized as follows: section II gives the brief overview of the related work. III covers the experimental results and analysis, and is subsequently followed by the conclusion.

## II. FEATURE LEVEL METHODS

Feature level methods are the next stage of processing where image fusion may take place. Fusion at the feature level requires extraction of features from the input images. Features can be pixel intensities or edge and texture features [11]. The Various kinds of features are considered depending on the nature of images and the application of the fused image. The features involve the extraction of feature primitives like edges, regions, shape, size, length or image segments, and features with similar intensity in the images to be fused from different types of images of the same geographic area. These features are then combined with the similar features present in the other input images through a pre-determined selection process to form the final fused image . The feature level fusion should be easy. However, feature level fusion is difficult to achieve when the feature sets are derived from different algorithms and data sources [12].

To explain the algorithms through this study, Pixel should have the same spatial resolution from two different sources that are manipulated to obtain the resultant image. So, before fusing two sources at a pixel level, it is necessary to perform a geometric registration and a radiometric adjustment of the images to one another. When images are obtained from sensors of different satellites as in the case of fusion of SPOT or IRS with Landsat, the registration accuracy is very important. But registration is not much of a problem with simultaneously acquired images as in the case of Ikonos/Quickbird PAN and MS images. The PAN images have a different spatial resolution from that of MS images. Therefore, resampling of MS images to the spatial resolution of PAN is an essential step in some fusion methods to bring the MS images to the same size of PAN, thus the resampled MS images will be noted by $M_k$ that represents the set of DN of band $k$ in the resampled MS image . Also the following notations will be used: P as DN for PAN image, $F_k$ the DN in final fusion result for band k. $\overline{M}_k$, $\overline{P}$ , and $\sigma_P, \sigma_{M_k}$ Denote the local means and standard deviation calculated inside the window of size (3, 3) for $M_k$ and P respectively.

### A. *Segment Based Image Fusion(SF):*

The segment based fusion was developed specifically for a spectral characteristics preserving image merge . It is based on an IHS transform [13] coupled with a spatial domain filtering. The principal idea behind a spectral characteristics preserving image fusion is that the high resolution of PAN image has to sharpen the MS image without adding new gray level information to its spectral components. An ideal fusion algorithm would enhance high frequency changes such as edges and high frequency gray level changes in an image without altering the MS components in homogeneous regions. To facilitate these demands, two prerequisites have to be addressed. First, color and spatial information have to be separated. Second, the spatial information content has to be manipulated in a way that allows adaptive enhancement of the images. The intensity $I_{LPF}$ of MS image is filtered with a low pass filter (LPF) [14-16] whereas the PAN image is filtered with an opposite high pass filter (HPF) [17-18].

HPF basically consists of an addition of spatial details, taken from a high-resolution Pan observation, into the low resolution MS image [19]. In this study, to extract the PAN channel high $P_{HPF}$ frequencies; a degraded or low-pass-filtered version of the PAN channel has to be created by applying the following set of filter weights (in a 3 x 3 convolution filter example) [14]:

$$LPF = \frac{1}{9}\begin{bmatrix} 1 & 1 & 1 \\ 1 & 1 & 1 \\ 1 & 1 & 1 \end{bmatrix} \quad (1)$$

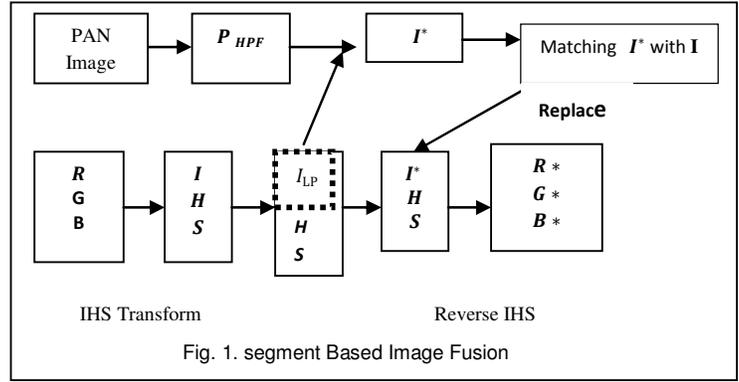

Fig. 1. segment Based Image Fusion

A LPF , which corresponds to computing a local average around each pixel in the image, is achieved. Since the goal of contrast enhancement is to increase the visibility of small detail in an image, subsequently, the (HPF) extracts the high frequencies using a subtraction procedure .This approach is known as Unsharp masking (USM) [20]:

$$P_{USM} = P - P_{LPF} \quad (2)$$

When this technique is applied, it really leads to the enhancement of all high spatial frequency detail in an image including edges, line and points of high gradient [21].

$$I^* = I_{LPF} + P_{UMS} \quad (3)$$

The low pass filtered intensity ($I_{LPF}$) of MS and the high pass filtered PAN band ($P_{UMS}$) are added and matched to the original intensity histogram. This study uses mean and standard deviation adjustment, which is also called adaptive contrast enhancement, as the following: [5]:

$$I^*{}_{New} = \overline{I} + (I^* - \overline{I}^*)\frac{\sigma_I}{\sigma_{I^*}} \quad (4)$$

σ and Mean adaptation are, in addition, a useful means of obtaining images of the same bit format (e.g., 8-bit) as the original MS image [22]. After filtering, the images are transformed back into the spatial domain with an inverse IHS and added together($I^*$) to form a fused intensity component with the low frequency information from the low resolution MS image and the high-frequency information from the high resolution PAN image. This new intensity component and the original hue and saturation components of the MS image form a new IHS image. As the last step, an inverse IHS transformation produces a fused RGB image that contains the spatial resolution of the panchromatic image and the spectral characteristics of the MS image. An overview flowchart of the segment Fusion is presented in Fig. 1.

### B. PCA-based Feature Fusion

The PCA is used extensively in remote sensing applications by many such as [23 -30]. It is used for dimensionality reduction, feature enhancement, and image fusion. The PCA is a statistical approach [27] that transforms a multivariate inter-correlated data set into a new un-correlated data set [31]. The PCA technique can also be found under the expression Karhunen Loeve approach [3]. PCA transforms or projects the features from the original domain to a new domain (known as PCA domain) where the features are arranged in the order of their variance. The features in the transformed domain are formed by the linear combination of the original features and are uncorrelated. Fusion process is achieved in the PCA domain by retaining only those features that contain a significant amount of information. The main idea behind PCA is to determine the features that explain as much of the total variation in the data as possible with as few of these features as possible. The PCA computation done on an N-by-N of MS image having 3 contiguous spectral bands is explained below. The computation of the PCA transformation matrix is based on the eigenvalue decomposition of the covariance matrix $\Sigma$ is defined as[33]:

$$\Sigma = \sum_{i=1}^{N^2}(\vec{X}_i - \vec{m})(\vec{X}_i - \vec{m})^T \quad (5)$$

where $\vec{X}_i$ is the $i^{th}$ spectral signature, $\vec{m}$ denotes the mean spectral signature and $N^2$ is the total number of spectral signatures. the total number of spectral signatures. In order to find the new orthogonal axes of the PCA space, Eigen decomposition of the covariance matrix $\Sigma$ is performed. The eigen decomposition of the covariance matrix is given by

$$\Sigma \vec{a}_k = \lambda_k \vec{a}_k \quad (6)$$

where $\lambda_k$ denotes the $k^{th}$ eigenvalue, $\vec{a}_k$ denotes the corresponding eigenvector and k varies from 1 to 3. The eigenvalues denote the amount of variance present in the corresponding eigenvectors. The eigenvectors form the axes of the PCA space, and they are orthogonal to each other. The eigenvalues are arranged in decreasing order of the variance. The PCA transformation matrix, A, is formed by choosing the eigenvectors corresponding to the largest eigenvalues. The PCA transformation matrix A is given by

$$A = [\vec{a}_1 | \vec{a}_2 | \dots | \vec{a}_J] \quad (7)$$

where $\vec{a}_1 \vec{a}_2 \dots \vec{a}_J$ are the eigenvectors associated with the J largest eigenvalues obtained from the eigen decomposition of the covariance matrix $\Sigma$. The data projected onto the corresponding eigenvectors form the reduced uncorrelated features that are used for further fusion processes.

Computation of the principal components can be presented with the following algorithm by [34]:
1. Calculate the covariance matrix $\Sigma$ from the input data.
2. Compute the eigenvalues and eigenvectors of $\Sigma$ and sort them in a descending order with respect to the eigenvalues.
3. Form the actual transition matrix by taking the predefined number of components (eigenvectors).
4. Finally, multiply the original feature space with the obtained transition matrix, which yields a lower-dimensional representation.

The PCA based feature fusion is shown in Fig. 2. The input MS are, first, transformed into the same number of uncorrelated principal components. Its most important steps are:
a. perform a principal component transformation to convert a set of MS bands (three or more bands) into a set of principal components.
b. Substitute the first principal component PC1 by the PAN band whose histogram has previously been matched with that of the first principal component. In this study the mean and standard deviation are matched by :

$$P_{New} = \bar{M}_K + (Pc1 - \overline{PC1})\frac{\sigma_M}{\sigma_{PC1}} \quad (8)$$

Perform a reverse principal component transformation to convert the replaced components back to the original image space. A set of fused MS bands is produced after the reverse transform [35-37].

The mathematical models of the forward and backward transformation of this method are described by [37], whose processes are represented by eq. (9) and (10). The transformation matrix contains the eigenvectors, ordered with respect to their Eigen values. It is orthogonal and determined either from the covariance matrix or the correlation matrix of the input MS. PCA performed using the covariance matrix is referred to as nun standardized PCA, while PCA performed using the correlation matrix is referred to as standardized PCA [37]:

$$\begin{bmatrix} PC1 \\ PC2 \\ \dots \\ PCn \end{bmatrix} = \begin{bmatrix} v11 & v21 & \dots & vn1 \\ v12 & v22 & \dots & vn2 \\ \dots & \dots & \dots & \dots \\ v1n & v2n & \dots & vnn \end{bmatrix} \begin{bmatrix} DN_{MS1}^l \\ DN_{MS2}^l \\ \dots \\ DN_{MSn}^l \end{bmatrix} \quad (9)$$

Where the transformation matrix

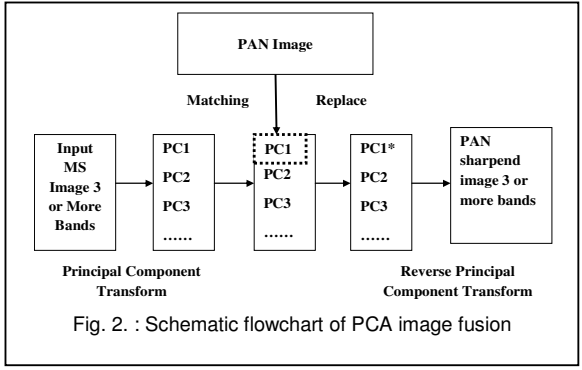

Fig. 2. : Schematic flowchart of PCA image fusion

$$v = \begin{bmatrix} v11 & v21 & \ldots & vn1 \\ v12 & v22 & \ldots & vn2 \\ \ldots & \ldots & \ldots & \ldots \\ v1n & v2n & \ldots & vnn \end{bmatrix}$$

$$\begin{bmatrix} DN_{MS1}^h \\ DN_{MS2}^h \\ \ldots \\ DN_{MSn}^h \end{bmatrix} = \begin{bmatrix} v11 & v21 & \ldots & vn1 \\ v12 & v22 & \ldots & vn2 \\ \ldots & \ldots & \ldots & \ldots \\ v1n & v2n & \ldots & vnn \end{bmatrix} \begin{bmatrix} DN_{PAN}^h \\ PC2 \\ \ldots \\ PCn \end{bmatrix} \quad (10)$$

(9) and (10) can be merged as follows:

$$\begin{bmatrix} DN_{MS1}^h \\ DN_{MS2}^h \\ \ldots \\ DN_{MSn}^h \end{bmatrix} = \begin{bmatrix} DN_{MS1}^l \\ DN_{MS2}^l \\ \ldots \\ DN_{MSn}^l \end{bmatrix} + (DN_{PAN}^{h'} - DN_{PAN}^l) \begin{bmatrix} v11 \\ v12 \\ \ldots \\ v1n \end{bmatrix} \quad (11)$$

Here $DN_{MS1,2,\ldots n}^l$ and $DN_{PAN}^h$ are the DN values of the pixels of different bands (1,2,….n) of MS and PAN images respectively and the superscripts $h$ and l denote high and low resolution. Also $DN_{PAN}^l$ = PC1 and $DN_{PAN}^{h'}$ is $DN_{PAN}^h$ stretched to have same mean and variance as PC1. The PCA based fusion is sensitive to the area to be sharpened because the variance of the pixel values and the correlation among the various bands differ depending on the land cover. So, the performance of PCA can vary with images having different correlation between the MS bands.

*C. Edge Fusion (EF):*

Edge detection is a fundamental tool in image processing and computer vision, particularly in the areas of feature detection and feature extraction, which aim at identifying points in a digital image at which the image brightness changes sharply or, more formally, has discontinuities [38]. The term 'edge' in this context refers to all changes in image signal value, also known as the image gradient [38].There are many methods for edge detection, but most of them can be grouped into two categories, search-based and zero-crossing based.

   a. **Edge detection based on First order difference – derivatives:**

usually a first-order derivative expression such as the gradient magnitude, and then searching for local directional maxima of the gradient magnitude using a computed estimate of the local orientation of the edge, usually the gradient direction, so the edge detection operator Roberts, Prewitt, Sobel returns a value for the first derivative in the horizontal direction ($M_y$) and the vertical direction ($M_x$). When applied to the PAN image the action of the horizontal edge-detector forms the difference between two horizontally adjacent points, as such detecting the vertical edges, Ex, as:

$$Ex_{x,y} = |P_{x,y} - P_{x+1,y}| \quad \forall x \in 1, N-1; y \in 1, N \quad (12)$$

To detect horizontal edges we need a vertical edge-detector which differences vertically adjacent points. This will determine horizontal intensity changes, but not vertical ones, so the vertical edge-detector detects the horizontal edges, Ey, according to:

$$Ey_{x,y} = |P_{x,y} - P_{x,y+1}| \quad \forall x \in 1, N; y \in 1, N-1 \quad (13)$$

Combining the two gives an operator E that can detect vertical and horizontal edges together.
That is,

$$E_{x,y} = |2 \times P_{x,y} - P_{x+1,y} - P_{x,y+1}| \, x,y \in 1, N-1 \quad (14)$$

This is equivalent to computing the first order difference delivered by Equation 14 at two adjacent points, as a new horizontal difference Exx, where

$$Exx_{x,y} = Ex_{x+1,y} + Ex_{x,y} = P_{x+1,y} - P_{x,y} + P_{x,y} - P_{x-1,y} = P_{x+1,y} - P_{x-1,y} \quad (15)$$

In this case the masks are extended to a $3*3$ neighbourhood, . The x and y masks given below are first convolved with the image to compute the values of and . Then the magnitude and angle of the edges are computed from these values and stored (usually) as two separate image frames. the edge magnitude, M, is the length of the vector and the edge direction, θ, is the angle of the vector:

$$M = Mx(x,y)^2 + My(x,y)^2 \quad (16)$$
$$\theta(x,y) = \tan^{-1}\left(\frac{M_y(x,y)}{M_x(x,y)}\right) \quad (17)$$

b. **Edge detection based on second-order derivatives:**
In 2-D setup, a commonly used operator based on second-order derivatives is the following Laplacian operator[39]:
$$\nabla^2 = \frac{\partial^2}{\partial x^2} + \frac{\partial^2}{\partial y^2} \quad (18)$$

For the image intensity function f(x, y), if a given pixel $(x_0, y_0)$ is on an edge segment, then $\nabla^2 f(x,y)$ has the zero-crossing properties around $(x_0, y_0)$ it would be positive on one side of the edge segment, negative on the other side, and zero at $(x_0, y_0)$ at someplace(s) between $(x_0, y_0)$ and its neighboring pixels [39].

The Laplace operator is a second order differential operator in the n-dimensional Euclidean space, There are many discrete versions of the Laplacian operator. The Laplacian mask used in this study as shown in Eq.20., by which only the given pixel and its four closest neighboring pixels in the z- and y-axis directions are involved in computation.

Discrete Laplace operator is often used in image processing e.g. in edge detection and motion estimation applications. Their extraction for the purposes of the proposed fusion can proceed according to two basic approaches: i) through direct edge fusion which may not result in complete segment boundaries and ii) through the full image segmentation process which divides the image into a finite number of distinct regions with discretely defined boundaries. In this instance. This study used first-order by using Sobel edge detection operator and second – order by discrete Laplacian edge detection operator as the following:

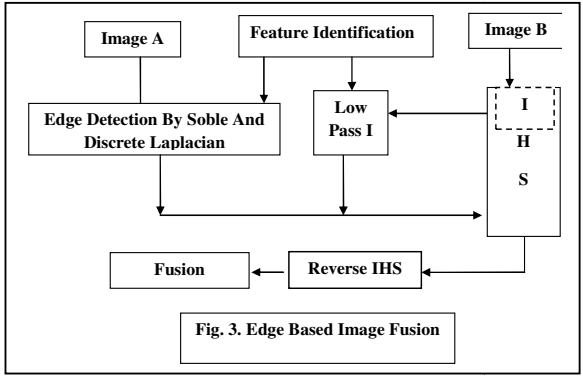

Fig. 3. Edge Based Image Fusion

- The Sobel operator was the most popular edge detection operator until the development of edge detection techniques with a theoretical basis. It proved popular because it gave, overall, a better performance than other contemporaneous edge detection operators, such as the Prewitt operator [40]. The templates for the Sobel operator as the following[41]:

$$G_x = \begin{bmatrix} -1 & -2 & -1 \\ 0 & 0 & 0 \\ 1 & 2 & 1 \end{bmatrix} \quad G_y = \begin{bmatrix} -1 & 0 & 1 \\ -2 & 0 & 2 \\ -1 & 0 & 1 \end{bmatrix} \quad (19)$$

- The discrete Laplacian edge detection operator

$$G_x = \begin{bmatrix} 0 & -1 & 0 \\ -1 & 5 & -1 \\ 0 & -1 & 0 \end{bmatrix} \quad G_y = \begin{bmatrix} 1 & -2 & 1 \\ -2 & 5 & -2 \\ 1 & -2 & 1 \end{bmatrix} \quad (20)$$

The proposed process of Edge Fusion is depicted in (Fig. 3) and consists of three steps:

1) edge detection of PAN image by soble and discrete Laplacian then subtraction the PAN from them.
2) Low Pas Filter the intensity of MS and add the edge of the pan.
3) After that, the images are transformed back into the spatial domain with an inverse IHS and added together($I^*$) to form a fused intensity component with the low frequency information from the low resolution MS image and the edge of PAN image. This new intensity component and the original hue and saturation components of the MS image form a new IHS image.
4) As the last step, an inverse IHS transformation produces a fused RGB image that contains the spatial resolution of the panchromatic image and the spectral characteristics of the MS image. An overview flowchart of the segment Fusion is presented in Fig. 3.

### III. EXPERIMENTS

In order to validate the theoretical analysis, the performance of the methods discussed above was further evaluated by experimentation. Data sets used for this study were collected by the Indian IRS-1C PAN (0.50 - 0.75 µm) of the 5.8 m resolution panchromatic band. Where the American Landsat (TM) the red (0.63 - 0.69 µm), green (0.52 - 0.60 µm) and blue (0.45 - 0.52 µm) bands of the 30 m resolution multispectral image were used in this **work**. Fig.4 shows the IRS-1C PAN and multispectral TM images. The scenes covered the same area of the Mausoleums of the Chinese Tang – Dynasty in the PR China [42] was selected as test sit in this study. Since this study is involved in evaluation of the effect of the various spatial, radiometric and spectral resolution for image fusion, an area contains both manmade and natural features is essential to study these effects. Hence, this work is an attempt to study the quality of the images fused from different sensors with various characteristics. The size of the PAN is 600 * 525 pixels at 6 bits per pixel and the size of the original multispectral is 120 * 105 pixels at 8 bits per pixel, but this is upsampled to by nearest neighbor. It was used to avoid spectral contamination caused by interpolation, which does not change the data file value. The pairs of images were geometrically registered to each other.

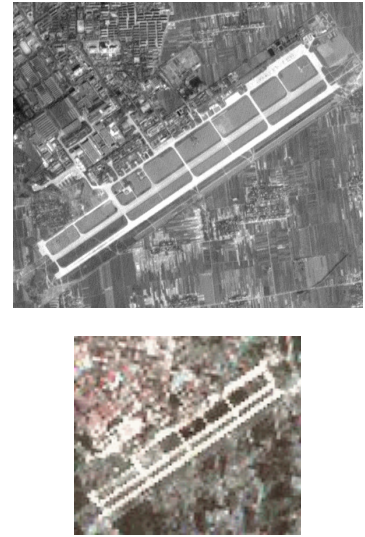

Fig.4. Original Panchromatic and Multispectral Images

To evaluate the ability of enhancing spatial details and preserving spectral information, some Indices including :

- Standard Deviation (SD),

$$\sigma = \sqrt{\frac{\sum_{i=1}^{m}\sum_{j=1}^{n}(BV(n,m)-\mu)^2}{m \times n}}$$

- Entropy(En) :in this study the measure of entropy values based on the first differences best than use the measure it based on the levels themselves.

$$En = -\sum_{0}^{I-1} P(i)\log_2 P(i)$$

- Signal-to Noise Ratio (SNR)

$$SNR_k = \sqrt{\frac{\sum_i^n \sum_j^m (F_k(i,j))^2}{\sum_i^n \sum_j^m (F_k(i,j)-M_k(i,j))^2}}$$

- Deviation Index (DI)

$$DI_k = \frac{1}{nm}\sum_i^n \sum_j^m \frac{|F_k(i,j) - M_k(i,j)|}{M_k(i,j)}$$

- Correlation Coefficient (CC)

$$CC = \frac{\sum_i^n \sum_j^m (F_k(i,j) - \overline{F}_k)(M_k(i,j) - \overline{M}_k)}{\sqrt{\sum_i^n \sum_j^m (F_k(i,j) - \overline{F}_k)^2}\sqrt{\sum_i^n \sum_j^m (M_k(i,j) - \overline{M}_k)^2}}$$

- Normalization Root Mean Square Error (NRMSE)

$$NRMSE_k = \sqrt{\frac{1}{nm * 255^2}\sum_i^n \sum_j^m (F_k(i,j) - M_k(i,j))^2}$$

where the $F_k$, $M_k$ are the measurements of each the brightness values of homogenous pixels of the result image and the original MS image of band k, $\overline{M}_k$ and $\overline{F}_k$ are the mean brightness values of both images and are of size $n * m$. BV is the brightness value of image data $\overline{M}_k$ and $\overline{F}_k$. To simplify the comparison of the different fusion methods, the values of the En, CC, SNR, NRMSE and DI index of the fused images are provided as chart in Fig. 5

## IV. DISCUSSION OF RESULTS

From table1 and Fig. 5 shows those parameters for the fused images using various methods. It can be seen that from Fig. 5a and table1 the SD results of the fused images remains constant for SF. According to the computation results En in table1, the increased En indicates the change in quantity of information content for radiometric resolution through the merging. From table1 and Fig.3b, it is obvious that En of the fused images have been changed when compared to the original multispectral excepted the PCA. In Fig.3c and table1 the maximum correlation values was for PCA , the maximum results of SNR was for SF. The results of SNR, NRMSE and DI appear changing significantly. It can be observed, from table1 with the diagram of Fig. 5d & Fig. 5e, that the results of SNR, NRMSE & DI of the fused image, show that the SF method gives the best results with respect to the other methods indicating that this method maintains most of information spectral content of the original MS data set which gets the same values presented the lowest value of the NRMSE and DI as well as the high of the CC and SNR. Hence, the spectral quality of fused image SF technique is much better than the others. In contrast, it can also be noted that the PCA image produce highly NRMSE & DI values indicating that these methods deteriorate spectral information content for the reference image. By combining the visual inspection results, it can be seen that the experimental results overall method are SF result which are the best results. Fig.6. shows the fused image results.

Table 1: Quantitative Analysis of Original MS and Fused Image Results Through the Different Methods

| Method | Band | SD | En | SNR | NRMSE | DI | CC |
|---|---|---|---|---|---|---|---|
| ORIGIN | 1 | 51.018 | 5.2093 | / | / | / | / |
|  | 2 | 51.477 | 5.2263 | / | / | / | / |
|  | 3 | 51.983 | 5.2326 | / | / | / | / |
| EF | 1 | 55.184 | 6.0196 | 6.531 | 0.095 | 0.138 | 0.896 |
|  | 2 | 55.792 | 6.0415 | 6.139 | 0.096 | 0.151 | 0.896 |
|  | 3 | 56.308 | 6.0423 | 5.81 | 0.097 | 0.165 | 0.898 |
| PCA | 1 | 47.875 | 5.1968 | 6.735 | 0.105 | 0.199 | 0.984 |
|  | 2 | 49.313 | 5.2485 | 6.277 | 0.108 | 0.222 | 0.985 |
|  | 3 | 47.875 | 5.1968 | 6.735 | 0.105 | 0.199 | 0.984 |
| SF | 1 | 51.603 | 5.687 | 9.221 | 0.067 | 0.09 | 0.944 |
|  | 2 | 52.207 | 5.7047 | 8.677 | 0.067 | 0.098 | 0.944 |
|  | 3 | 53.028 | 5.7123 | 8.144 | 0.068 | 0.108 | 0.945 |

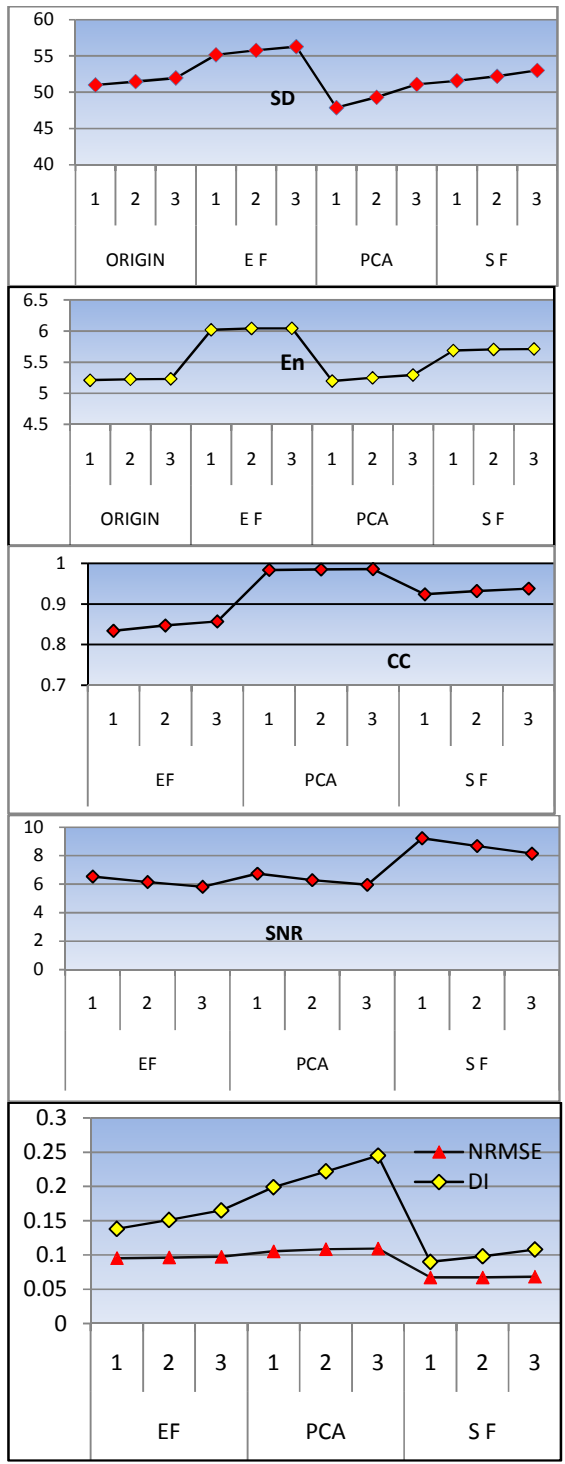

Fig. 5: Chart Representation of SD, En, CC, SNR, NRMSE & DI of Fused Images

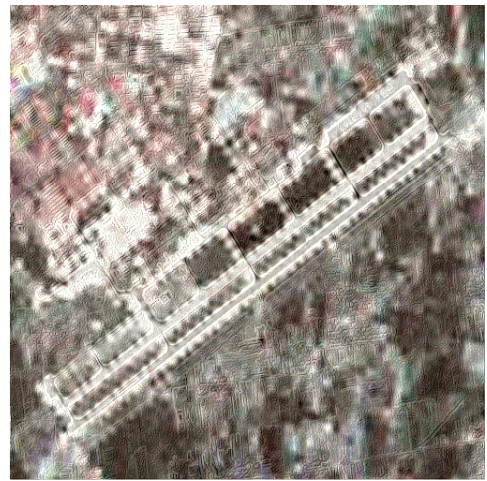

Edge Fusion

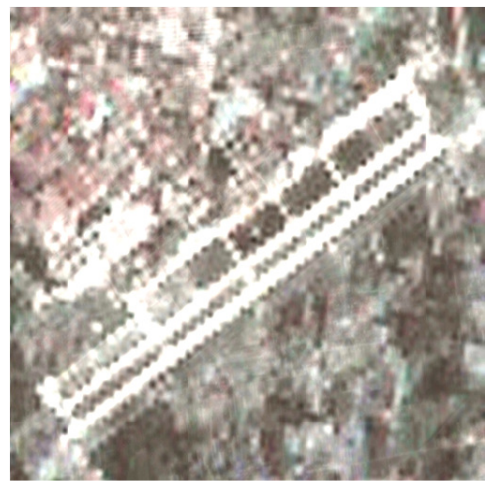

PCA

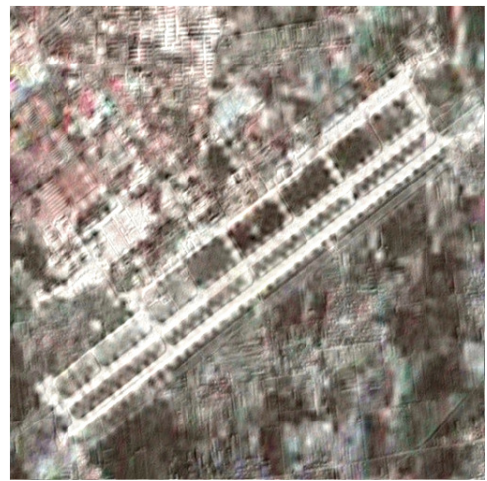

Segment Fusion

Fig.6: The Representation of Fused Images

## V. Conclusion

Image Fusion aims at the integration of disparate and complementary data to enhance the information apparent in the images as well as to increase the reliability of the interpretation. This leads to more accurate data and increased utility in application fields like segmentation and classification. In this paper, we proposed three different types of Image fusion techniques including PCA, SF and EF image fusion. Experimental results and statistical evaluation further show that the proposed SF technique maintains the spectral integrity and enhances the spatial quality of the imagery. the proposed SF technique yields best performance among all the fusion algorithms.

The use of the SF based fusion technique could, therefore, be strongly recommended if the goal of the merging is to achieve the best representation of the spectral information of MS image and the spatial details of a high-resolution PAN image. Also, the analytical technique of DI is much more useful for measuring the spectral distortion than NRMSE since the NRMSE gave the same results for some methods; but the DI gave the smallest different ratio between those methods, therefore , it is strongly recommended to use the DI because of its mathematical more precision as quality indicator.

**AUTHORS**

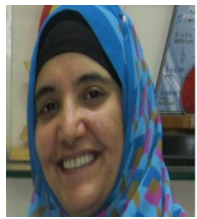

*Firouz Abdullah Al-Wassai*. Received the B.Sc. degree in, Physics from University of Sana'a, Yemen, Sana'a, in 1993. The M.Sc.degree in, Physics from Bagdad University , Iraqe, in 2003, Research student.Ph.D in the department of computer science (S.R.T.M.U), India, Nanded.

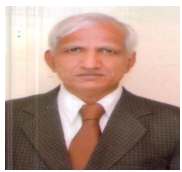

*Dr. N.V. Kalyankar*, Principal,Yeshwant Mahvidyalaya, Nanded(India) completed M.Sc.(Physics) from Dr. B.A.M.U, Aurangabad. In 1980 he joined as a leturer in department of physics at Yeshwant Mahavidyalaya, Nanded. In 1984 he completed his DHE. He completed his Ph.D. from Dr.B.A.M.U. Aurangabad in 1995. From 2003 he is working as a Principal to till date in Yeshwant Mahavidyalaya, Nanded. He is also research guide for Physics and Computer Science in S.R.T.M.U, Nanded. 03 research students are successfully awarded Ph.D in Computer Science under his guidance. 12 research students are successfully awarded M.Phil in Computer Science under his guidance He is also worked on various boides in S.R.T.M.U, Nanded. He is also worked on various bodies is S.R.T.M.U, Nanded. He also published 30 research papers in various international/national journals. He is peer team member of NAAC (National Assessment and Accreditation Council, India ). He published a book entilteld "DBMS concepts and programming in Foxpro". He also get various educational wards in which "Best Principal" award from S.R.T.M.U, Nanded in 2009 and "Best Teacher" award from Govt. of Maharashtra, India in 2010. He is life member of Indian "Fellowship of Linnean Society of London(F.L.S.)" on 11 National Congress, Kolkata (India). He is also honored with November 2009.

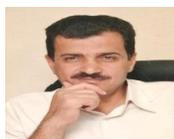

*Dr. Ali A. Al-Zuky*. B.Sc Physics Mustansiriyah University, Baghdad , Iraq, 1990. M Sc. In1993 and Ph. D. in1998 from University of Baghdad, Iraq. He was supervision for 40 postgraduate students (MSc. & Ph.D.) in different fields (physics, computers and Computer Engineering and Medical Physics). He has More than 60 scientific papers published in scientific journals in several scientific conferences.